\pdfoutput=1

\documentclass[11pt]{article}

\usepackage{ACL2023}

\usepackage{times}
\usepackage{latexsym}
\usepackage{graphicx,xcolor}  
\usepackage{pxrubrica}        
\usepackage{url}
\usepackage{subcaption}
\usepackage{caption}
\usepackage{amsmath}
\usepackage{multirow}
\usepackage{siunitx}
\usepackage{bm}
\usepackage{booktabs}

\usepackage[T1]{fontenc}

\usepackage[utf8]{inputenc}

\usepackage{microtype}

\usepackage{inconsolata}

\title{Likelihood-based Mitigation of Evaluation Bias in Large Language Models}

\author{%
Masanari Oi ${}^{1}$ Masahiro Kaneko ${}^{2, 1}$ Ryuto Koike ${}^{1}$ Mengsay Loem ${}^{1}$ Naoaki Okazaki ${}^{1}$
\\ ${}^{1}$ Tokyo Institute of Technology ${}^{2}$ MBZUAI
\\ \texttt{\{masanari.ohi@nlp., ryuto.koike@nlp., mengsay.loem@nlp., okazaki@\}c.titech.ac.jp} \\
\texttt{masahiro.kaneko@mbzuai.ac.ae}
}

\begin{document}
\maketitle
\begin{abstract}
Large Language Models (LLMs) are widely used to evaluate natural language generation tasks as automated metrics.
However, the likelihood, a measure of LLM's plausibility for a sentence, can vary due to superficial differences in sentences, such as word order and sentence structure.
It is therefore possible that there might be a \textbf{likelihood bias} if LLMs are used for evaluation: they might overrate sentences with higher likelihoods while underrating those with lower likelihoods.
In this paper, we investigate the presence and impact of likelihood bias in LLM-based evaluators.
We also propose a method to mitigate the likelihood bias.
Our method utilizes highly biased instances as few-shot examples for in-context learning.
Our experiments in evaluating the data-to-text and grammatical error correction tasks reveal that several LLMs we test display a likelihood bias.
Furthermore, our proposed method successfully mitigates this bias, also improving evaluation performance (in terms of correlation of models with human scores) significantly.
\end{abstract}

\section{Introduction}
Large Language Models (LLMs) exhibit robust language comprehension and text generation capabilities~\cite{anil2023palm, openai2023gpt4}.
Relying on this ability, recent studies~\cite{liu-etal-2023-g, kocmi-federmann-2023-large, chiang-lee-2023-large} have employed LLMs as evaluators for natural language generation tasks, surpassing the performance of existing automatic evaluation methods such as BLEU~\cite{papineni-etal-2002-bleu} and ROUGE~\cite{lin-2004-rouge}.
To assess the quality of a text, LLMs output a comprehensive evaluation score based on criteria such as fluency and meaning.

\begin{figure}[t]
    \centering
    \includegraphics[width=1.0\columnwidth]{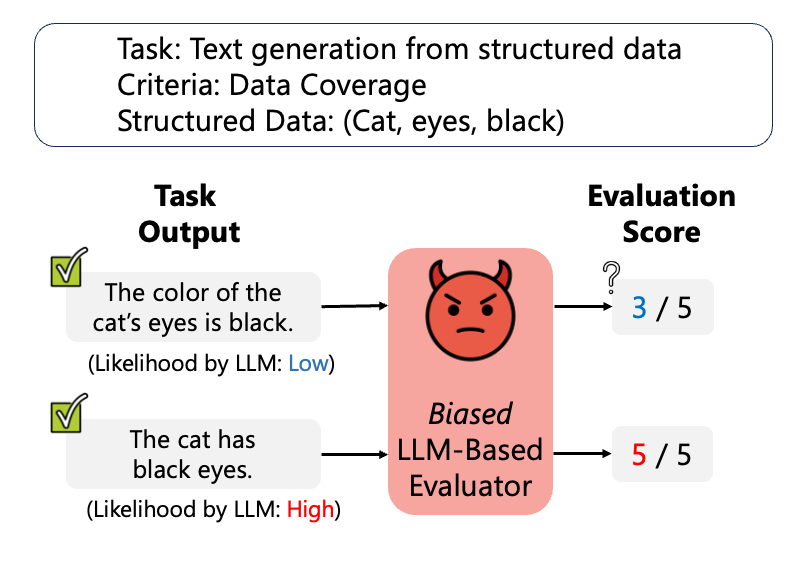}
    \caption{An example of likelihood bias. Correct, but low-likelihood output (top) is scored low while high-likelihood output (bottom) is scored high.}
    \label{fig:bias_image}
\end{figure}

LLMs generate a text based on the likelihood estimations derived from the training process that aims to maximize the likelihood of their large-scale training data.
Consequently, it is intuitively possible that the likelihood of a text influences the generation of its evaluation score.
However, the likelihood estimations by LLMs may not necessarily align with the quality of the text.
For instance, the likelihood calculated by the LLM fluctuates due to superficial differences in sentences, such as word order and sentence structure, even for sentences with identical meaning~\cite{kuribayashi-etal-2020-language}.
Such fluctuation of likelihood could negatively impact the evaluation score based on meaning criteria.

In this paper, we introduce \textbf{likelihood bias}, where LLM-based evaluators overrate high-likelihood sentences and underrate low-likelihood ones compared to human scores.
Figure \ref{fig:bias_image} shows one example of likelihood bias.
Here, a biased evaluator gives a lower score of 3/5 to a correct but low-likelihood output (top) while giving a higher score of 5/5 to a high-likelihood output (bottom), based on the criteria of data coverage.
Addressing this issue, we propose a method that a) quantifies and b) mitigates likelihood bias.
We quantify the bias by using the correlation between the disparity in evaluation scores generated by LLMs and those provided by human evaluators, and the likelihood of a target text.
Our bias reduction method identifies and utilizes highly biased instances as few-shot examples for in-context learning.

The extent of likelihood bias may vary with the evaluation criteria used.
For instance, likelihood bias is anticipated to be more pronounced in criteria like data coverage, which is less directly related to likelihood.
Conversely, the bias is expected to be less significant in criteria like fluency, which is closely related to likelihood.
To verify this characteristic of likelihood bias, we adopt two tasks: data-to-text and Grammatical Error Correction (GEC).
We use these tasks because, unlike most existing data~\cite{freitag-etal-2021-experts, guan-etal-2021-openmeva, kamalloo-etal-2023-evaluating}, the evaluation data for these tasks include multiple criteria such as fluency, grammar, and data coverage.

Our experimental results show that both evaluators based on GPT-3.5 and Llama2-13B~\cite{touvron2023llama} indeed suffer from likelihood bias.
Moreover, our bias reduction method mitigates likelihood bias, and improves evaluation performance in many cases.
For reproducibility, we release our code~\footnote{\url{https://github.com/stjohn2007/likelihood_bias}}.

\section{Method}
Following a common methodology in LLM-based evaluation~\cite{liu-etal-2023-g, chiang-lee-2023-large}, we calculate the LLM's evaluation score $\text{Score}_{\text{m}}$ based on the models' response to a prompt.
Specifically, we calculate $\text{Score}_{\text{m}}$ as the expected value over candidate scores (e.g. \{1, 2, 3, 4, 5\}) based on the probability that models output these scores, following the setting of \citet{liu-etal-2023-g}.
Our prompt includes a task description and the evaluation criteria, and several few-shot example instances for in-context learning~\footnote{The actual prompts and exact equation we use to calculate the $\text{Score}_{\text{m}}$ are provided in Appendix \ref{sec:app_eval}.}.
The reason we use in-context learning is that it is known to stabilize the model.
This puts us in a position to quantify the strength of likelihood bias. 

\subsection{Measuring Likelihood Bias}\label{sec:method_measure}
We define \textbf{likelihood bias} in LLM-based evaluators as the tendency to overrate high-likelihood sentences and underrate low-likelihood ones, compared to human ratings.
First, we calculate LS, the \textbf{Likelihood Score}, representing the likelihood $P$ calculated by LLM.
Given an instance $t$ with input $t_{i}$, output $t_{o}$, task description $d$, and model parameters $\theta$, LS is defined as follows:

\begin{equation}
    \text{LS}(t) = \log P(t_{o} \mid t_{i}, d ; \theta) \label{eq:LS}
\end{equation}

We next calculate US, \textbf{Unfairness Score}, which represents the difference between scores by LLM 
($\text{Score}_{\text{m}}$) and scores by humans ($\text{Score}_{\text{h}}$).
To account for different scoring ranges between models and humans, $\text{Score}_{\text{m}}$ and $\text{Score}_{\text{h}}$ are normalized so that they have the same mean and range.
\begin{equation}
    \text{US}(t) = \text{Score}_{\text{m}}(t ; \theta) - \text{Score}_{\text{h}}(t) \label{eq:US}
\end{equation}
When measuring the bias, we choose eight few-shot example instances at random.

\textbf{BiasScore} is then our metric that measures likelihood bias, which is calculated as the correlation in terms of Spearman's rank correlation coefficient~$\rho$ between Likelihood Score and Unfairness Score across a Dataset $D = \{t^{(1)}, t^{(2)}, \ldots, t^{(n)} \}$, using each instance $t^{(i)}$:
\begin{gather}
    \text{LS}_{\text{D}} = \lbrack \text{LS}(t^{(1)}), \text{LS}(t^{(2)}), \ldots, \text{LS}(t^{(n)}) \rbrack \\
    \text{US}_{\text{D}} = \lbrack \text{US}(t^{(1)}), \text{US}(t^{(2)}), \ldots, \text{US}(t^{(n)}) \rbrack \\
    \text{BiasScore} = \rho (\text{LS}_{\text{D}}, \text{US}_{\text{D}}) \label{eq:BS}
\end{gather}
BiasScore ranges from -1 to 1, where 1 indicates strong likelihood bias, -1 implies the opposite bias from what we assume, and 0 suggests no bias.

\begin{figure}[t]
    \centering
    \includegraphics[width=1.0\columnwidth]{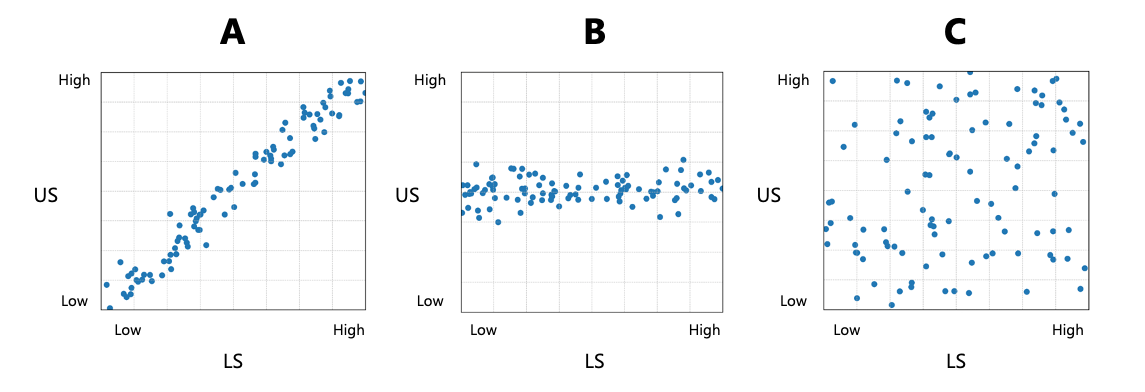}
    \caption{Likelihood bias of hypothetical evaluators. A: biased, B: unbiased with high performance, and C: unbiased with low performance.}
    \label{fig:bias_graph}
\end{figure}

\subsection{Mitigating Likelihood Bias}\label{sec:method_mitigate}

Figure~\ref{fig:bias_graph} plots LS against US in order to show the likelihood bias of multiple hypothetical evaluators~\footnote{Please note that the figure is the pseudo-scatter plot that represents hypothetical evaluators. We do not include concrete values for LS and US in the figure since LS does not have the lower bound, and the scale of US depends on the dataset.}.
Each point represents a pair of scores for an instance.
The BiasScore corresponds to the slope of the main cluster of instances. 

Figure~\ref{fig:bias_graph} (A) shows a middle-performing and biased evaluator. It unfairly gives high ratings to texts with high likelihood (points in the upper right) and low ratings to texts with low likelihood (points in the lower left).
We assume that LLM-based evaluators are in this state before bias mitigation.
Figure~\ref{fig:bias_graph} (B) shows the ideal outcome of mitigation: the BiasScore is zero (i.e., there is no bias), and the performance remains high. 
There is also no bias in Figure~\ref{fig:bias_graph} (C) (and thus BiasScore = 0), but this evaluator is of no use as the output is random (low-performance). 

The target of our bias mitigation strategy is to change situation (A) into (B), while avoiding low evaluation performance as in (C).
We concentrate on highly biased instances (top-right and bottom-left points in (A)) in our training data. 
For this, we require an instance-based measure of bias, which is provided by $\text{RS}(t)$ as follows:
\begin{gather}
     \text{RS}(t) = |\text{LS}^*(t) + \text{US}^*(t)|
\end{gather}
Here, $\text{LS}^*$ and $\text{US}^*$ are normalized so that they both have an average of 0 and a range from -1 to 1 across a dataset $D$.
$\text{RS}(t)$ is high for instances $t$ that are closer to the top-right or bottom-left of the scatter plot. 
For our mitigation strategy, we choose instances with the highest RS(t) from the training data, and use these instances as few-shot examples for in-context learning, after replacing the LLM scores with the human gold-standard scores.

\section{Experiments}

\subsection{Settings}
\paragraph{Datasets}
From a limited set of tasks with available datasets assessed by humans on multiple criteria, we selected two for our experiments: a) data-to-text, converting RDF data into English sentences, and b) GEC.
For data-to-text, we use WebNLG+~\cite{castro-ferreira-etal-2020-2020}, which contains 2846 instances.
$\text{Score}_{\text{h}}$ is provided by human judges, who rated each instance on five criteria (text structure, relevance, fluency, correctness and data coverage).
For GEC, we use the TMU-GFM-Dataset~\cite{yoshimura-etal-2020-reference}, which contains 4221 instances.
$\text{Score}_{\text{h}}$ is provided by human judges, who rated each instance on two criteria (grammar and fluency~\footnote{All criteria and their definitions are given in Appendix \ref{sec:app_dataset}. The original GEC dataset contains a third criterion, meaning. However, we exclude this criterion because it does not contribute to the overall evaluation~\citep{yoshimura-etal-2020-reference}.}).
We split each dataset into training and test data at a ratio of 4:1.

\paragraph{Models}
The  LLMs used in our experiments are GPT-3.5 provided via API by OpenAI~\footnote{We use \texttt{gpt-3.5-turbo-instruct} as the model in API call.} and Llama2-13B (L-13B).
For GPT-3.5, since it does not support the output of token generation likelihood, we use Llama2-13B's likelihood as an approximation.
We first measure how well the LLMs work as evaluators, using Spearman's rank correlation coefficient $\rho$ between human and model scores. 
The ``Before'' column of Evaluation Performance in Table~\ref{tab:result} shows these results.
The ballpark figures are that GPT-3.5 is the superior system for data-to-text, while for GEC, it roughly performs on a par with Llama2-13B. 

\subsection{Measuring Likelihood Bias} \label{sec:measure}

\begin{table*}[t]
\centering
\small
\tabcolsep 3pt
\begin{tabular}{llllllllll}
\toprule
\ & \     & \multicolumn{4}{c}{BiasScore} & \multicolumn{4}{c}{Evaluation Performance $\rho$} \\
\cmidrule(lr){3-6} \cmidrule(lr){7-10}
\ & \     & \multicolumn{2}{c}{Before} & \multicolumn{2}{c}{After} & \multicolumn{2}{c}{Before} & \multicolumn{2}{c}{After} \\
\cmidrule(lr){3-4} \cmidrule(lr){5-6} \cmidrule(lr){7-8} \cmidrule(lr){9-10}
Task & Criterion & GPT-3.5 & L-13B & GPT-3.5 & L-13B & GPT-3.5 & L-13B & GPT-3.5 & L-13B \\
\midrule
\multirow{7}{*}{D2T} & text structure & .36 & .17 & \bf{.23}~* & \bf{.02}~* & .46 & .34 & \bf{.53}~* & \bf{.36} \\
& relevance & .43 & .28 & \bf{.31}~* & \bf{.15}$~^{\dag}$ & .35 & .25 & \bf{.38} & .23 \\
& fluency & .26 & .20 & .29 & \bf{.00}~* & .41 & .33 & \bf{.55}~* & \bf{.52}$~^{\dag}$ \\
& correctness & .36 & .21 & \bf{.32} & \bf{-.01}~* & .44 & .37 & \bf{.47} & \bf{.43} \\
& data coverage & .40 & .24 & \bf{.32}~* & .16 & .20 & .24 & \bf{.30}$~^{\dag}$ & \bf{.25} \\
\cmidrule(lr){2-10}
& total (micro) & .38 & .17 & \bf{.32}$~^{\dag}$ & \bf{.02}$~^{\dag}$ & .48 & .40 & \bf{.58}~* & \bf{.46} \\
\cmidrule(lr){1-10}
\multirow{4}{*}{GEC} & grammar & .46 & .24 & \bf{.37}$~^{\dag}$ & .24 & .48 & .45 & \bf{.54} & \bf{.46} \\
& fluency & .36 & .16 & \bf{.29} & \bf{.09} & .40 & .49 & \bf{.47} & .48 \\
\cmidrule(lr){2-10}
& total (micro) & .43 & .21 & \bf{.37} & \bf{.18} & .45 & .48 & \bf{.52} & \bf{.52} \\
\bottomrule
\end{tabular}
\caption{BiasScore and Evaluation performance before and after mitigating likelihood bias. Values affected positively by our mitigation method appear boldfaced. *~represents significant difference ( $p < 0.05$ ) between before and after mitigation. \dag ~ represents marginal significant difference ( $p < 0.06$ ).}
\label{tab:result}
\end{table*}

We use the method described in Section \ref{sec:method_measure} for likelihood bias measurement.
We introduce a new criterion representing the overall result, total, by micro-averaging over the criteria~\footnote{Please note that when micro-averaging, the BiasScores reported in Table~\ref{tab:result} is not an average of the BiasScores of the individual evaluation criteria, since to calculate the total BiasScore we first average over the human and LLM evaluation scores and then apply Equation \ref{eq:BS}.}.

\paragraph{Results for data-to-text}
The ``Before'' column of the ``D2T'' row of BiasScores in Table~\ref{tab:result} reveals a bias for both models, with BiasScore for most evaluation criteria exceeding 0.17.
Across all criteria (total), GPT-3.5 has the strongest bias (0.38), followed by Llama2-13B (0.17).
Relevance is the criterion with the strongest bias in both models, GPT-3.5 (0.43) and Llama2-13B (0.28).

\paragraph{Results for GEC}
The ``Before'' column of the ``GEC'' row of BiasScores in Table~\ref{tab:result} also shows bias in both models and evaluation criteria: all BiasScores exceed 0.16.
As with data-to-text, GPT-3.5 overall displays a stronger bias across all criteria (0.43) than Llama2-13B (0.21).

\paragraph{Intrinsic vs non-intrinsic evaluation criteria}
Looking at the ``Before'' column of the ``D2T'' row of BiasScores in Table~\ref{tab:result}, there are two evaluation criteria which display relatively small likelihood biases across both models, namely fluency and text structure.
These criteria are concerned with text quality alone and they are intrinsic to the output text.
The criteria are true of the output text to a higher or lesser degree, but this is independent of what the input looked like. 
In contrast, relevance and data coverage are dependent on external factors in the input. 
The quality definition for those criteria is affected by the process that transforms the input into the output. 
Therefore, such criteria are not intrinsic.
From our results, we see that there is a marked difference in BiasScore between non-intrinsic and intrinsic criteria: non-intrinsic criteria are much more prone to bias.
These results suggest an intuitive interpretation:
The effect of the likelihood on the evaluation does not necessarily cause a harmful bias on intrinsic criteria as much as on non-intrinsic ones.
This might be because likelihood is intrinsic to the output text, and thus, likelihood is strongly related to intrinsic criteria~\footnote{We provide the reason we don't focus on the criteria of GEC and discuss the criterion of correctness in intrinsic / non-intrinsic paradigm in Appendix~\ref{sec:app_criteria}.}.

\subsection{Mitigating Likelihood Bias} \label{sec:mitigate}
We now use the method described in Section~\ref{sec:method_mitigate}, with eight highly biased examples for mitigation.
Notably, as stated in Section~\ref{sec:method_measure}, we also employ eight randomly picked examples for in-context learning when measuring bias, meaning the difference between before and after mitigation is only how we choose few-shot example instances.
In the ``After'' columns of Table~\ref{tab:result}, we boldface the value if our method brings a BiasScore close to zero or if it improves evaluation performance.
We test for the significance of differences using the two-sided randomized pair-wise permutation test with R=100000 and $\alpha$=0.05.
If a difference between unmitigated and mitigated conditions is significant, we indicate this with an asterisk (*); marginal significance ($p<0.06$) is indicated using a dagger (\dag).

\paragraph{Results in data-to-text}
The ``After'' column of the ``D2T'' row of BiasScores and Evaluation performance in Table~\ref{tab:result} shows that our method brings the BiasScore closer to zero and increases evaluation performance across the board.
With our method, the BiasScores decrease significantly for Llama2-13B for text structure (-0.15), fluency (-0.20), and correctness (-0.20). 
For GPT-3.5, results are significantly decreased for text structure (-0.13), relevance (-0.12), and data coverage (-0.08).
At the same time, the evaluation performance improves significantly for GPT-3.5 by +0.10 for total, by +0.14 for fluency, with marginally significant differences for GPT-3.5 in text structure, data coverage. 
For Llama2-13B, the only criterion with a marginally significant improvement is fluency.
We consider this an overall successful mitigation. 

\paragraph{Results for GEC}
As with data-to-text, the ``After'' column of the ``GEC'' row of BiasScores and Evaluation performance in Table~\ref{tab:result} shows our method brings the BiasScore closer to zero and improves evaluation performance in many cases.
Although few criteria achieve significant differences either in BiasScore or evaluation performance, our method at least shows changes in the right direction.

In summary, the results for the data-to-text and GEC tasks imply that our mitigation strategy can decrease the likelihood bias of LLMs and improve the evaluation performance simultaneously~\footnote{We conduct further experiments on visualization and case study about the mitigation of bias in Appendix \ref{sec:app_vis}.}.

\section{Related Work}
\paragraph{LLM-based evaluator}
\citet{kocmi-federmann-2023-large} employ nine GPT models to evaluate translation tasks and verify that GPT-3.5 and larger models possess sufficient capabilities for task evaluation.
\citet{chiang-lee-2023-large} show that several LLMs' evaluation results are consistent with those obtained by expert human evaluation in open-ended story generation and adversarial attacks. They also find that LLM-based evaluators are stable over different prompts and sampling algorithms.
\citet{NEURIPS2023_91f18a12} employ GPT-4 to evaluate the conversation ability of LLMs on two benchmarks: MT-bench and Chatbot Arena.
\citet{liu-etal-2023-g} propose G-EVAL, a framework of LLM-based evaluation. They incorporate chain-of-thoughts (CoT)~\cite{wei2023chainofthought} and introduce a form-filling paradigm and scoring method for stable and fine-grained evaluation.
Although LLM-based evaluators have been employed to evaluate several tasks, such as translation~\cite{kocmi-etal-2021-ship}, summarization~\cite{liu-etal-2023-g}, story generation~\cite{chiang-lee-2023-large}, multi-turn conversation~\cite{NEURIPS2023_91f18a12}, in this work, we focus on two tasks, data-to-text and GEC, to inspect likelihood bias on multiple criteria.

\paragraph{Biases in LLM-based evaluators}
\citet{NEURIPS2023_91f18a12} define \textit{self-enhancement bias} as the tendency of LLM-based evaluators to favor the answers generated by themselves. Their preliminary experiments indicate the existence of this bias. A similar tendency is also reported by \citet{liu-etal-2023-g}.
Also, \citet{NEURIPS2023_91f18a12} introduce \textit{verbosity bias}, referring to the tendency of LLM-based evaluators to favor longer, more verbose responses. \citet{saito2023verbosity} propose a metric to measure verbosity bias and find both GPT-4 and GPT-3.5 exhibit this bias according to their metric.
In contrast to these findings, we introduce \textit{likelihood bias}, where LLM-based evaluators evaluate high-likelihood sentences and underrate low-likelihood ones compared to human scores. Notably, this is the first work simultaneously proposing methodologies to quantify and mitigate a bias in LLM-based evaluators.

\section{Conclusion} \label{sec:con}
This paper identifies likelihood bias in LLMs as the phenomenon of LLMs overrating high-likelihood texts and underrating low-likelihood ones. 
We introduce a method for quantifying bias and propose a solution to the bias problem: using highly biased instances as few-shot examples for in-context learning.
Experiments with two tasks (data-to-text and GEC) show that LLMs exhibit strong likelihood bias, and that our method successfully mitigates it, improving evaluation performance.

In future work, we aim to investigate the relationship between other biases in LLM-based evaluators, such as self-enhancement bias~\cite{NEURIPS2023_91f18a12} and verbosity bias~\cite{NEURIPS2023_91f18a12, saito2023verbosity}, and likelihood bias. For instance, we may be able to explain self-enhancement bias through the concept of likelihood bias, as the answers generated by LLMs might have high likelihoods calculated by the models themselves.
Additionally, we plan to examine the impact of instruction-tuning and model size on likelihood bias.

\newpage
\section*{Limitations}
Our work has several limitations. 
($\mathrm{i}$)
Since our method uses in-context learning, the number of tokens that can be used is limited.
Therefore, our method may not be suitable for tasks with long input or output lengths, such as summarization, as the amount of space that can be used is even more limited. 
($\mathrm{ii}$)
In-context learning also brings another limitation.
Since it increases the prompt length, the computational (or API call) costs also go up compared to a zero-shot setting.
Again, please note that these limitations are derived from in-context learning, and our method doesn't increase prompt length and degrade efficiency compared to the settings that employ in-context learning.
One solution to them is fine-tuning the model instead of in-context learning.
It is therefore necessary to explore whether fine-tuning works better than in-context learning and how much data we need. 

\section*{Ethics Statement}
While we do not foresee any ethical risks caused by our research, LLMs not only exhibit biased likelihood based on surface-level information such as words and sentence structure but also on information like gender, religion, and race~\cite{kaneko2023impact,oba2023contextual,anantaprayoon2023evaluating}.
For instance, LLMs might assign a higher likelihood to \textit{``She is a nurse''} compared to \textit{``He is a nurse''}.
Reducing likelihood bias could potentially address social bias in evaluators.
However, it is worth noting that this work does not investigate such aspects, and this remains a task for future research.

\section*{Acknowledgements}
These research results were obtained from the commissioned research (No.22501) by National Institute of Information and Communications Technology (NICT) , Japan.
We are grateful to Simone Teufel, a professor at the University of Cambridge, for her valuable advice and feedback.

\newpage

\bibliography{anthology,custom}
\bibliographystyle{acl_natbib}

\newpage
\appendix

\section{LLM evaluation method}
\label{sec:app_eval}

\paragraph{Calculation of likelihood}
As shown in Equation \ref{eq:LS}, we calculate the likelihood of task output $t_{o}$ based on task description $d$ and task input $t_{i}$.
This approach aims to obtain a more contextually relevant likelihood, factoring in both the specifics of the task and the input, rather than simply calculating $\log{P(t_{o} ; \theta)}$.
Specific examples of task description $d$ are indicated below.
\begin{itemize}
    \item data-to-text: \textit{Please generate a description of the following xml data}
    \item GEC: \textit{Please modify the following English text to make it grammatically correct}
\end{itemize}

\paragraph{Calculation of $\text{Score}_{\text{m}}$}
As is common in LLM-based evaluation~\cite{liu-etal-2023-g, chiang-lee-2023-large}, the model is given a prompt $I$, which includes a task description, the evaluation criteria, and an instance $t$, and then predicts score $\text{Score}_{m}$.
We also use in-context learning, with the intention of stabilizing the model.
Examples are chosen at random when measuring the bias, and are chosen according to the method described in Section \ref{sec:method_mitigate} when mitigating the bias.
Finally, we calculate $\text{Score}_{\text{m}}$ as the expected score over scores.
We follow the setting of~\citet{liu-etal-2023-g}, who have observed that using the expected score, considering the model's distribution over scores for each instance, rather than always taking the most likely score, leads to a more robust evaluation. 
Given score candidates $\{ 1, 2, ..., n \}$, the probability of each score $Q(i \mid t, F, I; \theta)$,  $\text{Score}_{\text{m}}$ is formulated as follows:
\begin{equation}
    \text{Score}_{\text{m}} (t ; \theta) = \frac{\sum_{i = 1}^{n} i \times Q(i \mid t, F, I ; \theta)} {\sum_{j = 1}^{n} Q(j \mid t, F, I ; \theta)}
\end{equation}

\paragraph{Example Prompts}
Here, we provide two examples of the prompts used for LLM-based evaluators.
Our prompts are inspired by the prompts \citet{liu-etal-2023-g} used.

\textbf{Evaluate Correctness in data-to-text}
\begin{quote}
    You will be given an xml data and an English sentence that represents xml data.
    Your task is to rate the sentence that represents xml data on one metric.
    Please make sure you read and understand these instructions carefully. Please keep this document open while reviewing, and refer to it as needed.
    Evaluation Criteria:
    Correctness: (1-5) - does the text describe predicates with correct objects and does it introduce the subject correctly? 1 is the lowest score, 5 is the highest.
\end{quote}

\textbf{Evaluate Fluency in GEC}
\begin{quote}
    You will be given an English sentence that may have grammatical errors and a sentence that is the corrected version of the sentence.
    Your task is to rate the corrected sentence on one metric.
    Please make sure you read and understand these instructions carefully. Please keep this document open while reviewing, and refer to it as needed.
    Evaluation Criteria:
    Fluency: (0-4) -  How natural the sentence sounds for native speakers; 4: Extremely natural, 3: Somewhat natural, 2: Somewhat unnatural, and 1: Extremely unnatural, and 0: Other.
\end{quote}

\section{Dataset} \label{sec:app_dataset}
\paragraph{data-to-text}
We use WebNLG+ Dataset (CC BY-NC-SA 4.0) ~\cite{castro-ferreira-etal-2020-2020}.
Specifically, we collect instances that have human evaluation scores from their dataset.
The total number of instances we use is 2846.
We use them following their license.
There are five criteria in the original dataset:
\begin{itemize}
    \item text structure: whether the output is grammatically correct and well-structured
    \item relevance: whether the output is based on the input information
    \item fluency: whether the output is natural
    \item correctness: whether the output explains the input data correctly and reasonably
    \item data coverage: whether the output includes all the input data
\end{itemize}
Human annotators rate each instance on these criteria using a 100-point scale from 0 to 100.

\paragraph{GEC}
We use the TMU-GFM-Dataset (CC BY 4.0) ~\cite{yoshimura-etal-2020-reference}, which contains 4221 instances.
We use them following their license.
There are three criteria in the original dataset:
\begin{itemize}
    \item grammar: whether the output is grammatically correct
    \item fluency: whether the output is natural
    \item meaning: whether the output has the same meaning as the input
\end{itemize}
Human annotators rate each instance on these criteria using a 5-point scale from 0 to 4.
As mentioned in the footnote, we exclude meaning because, according to the original paper \cite{yoshimura-etal-2020-reference}, it does not contribute to the overall evaluation.

\section{Hyperparameters}
To guarantee reproducibility as much as possible, we set the hyperparameters on API calls to make GPT-3.5 deterministic.
We use \texttt{temperature} of 0, \texttt{top\_p} of 0.

As for the number of few-shot examples for in-context learning, we use eight examples.
This is the reasonable value that models can learn several pieces of information without violating the limit on the number of input tokens.

\section{Computational Budget}
We run all the experiments on ABCI (\url{https://abci.ai/}), Compute Node(A), whose CPUs are two Intel Xeon Platinum 8360Y, and GPUs are eight NVIDIA A100 SXM4.
The approximate total processing time is 30 hours.

\section{Additional Discussion on intrinsic/non-intrinsic evaluation criteria} \label{sec:app_criteria}
We do not focus on the criteria of GEC within the intrinsic/non-intrinsic paradigm in Section~\ref{sec:measure}, as its criteria, fluency and grammar, are both intrinsic.
We also do not include a discussion of correctness within the paradigm because it exhibits a relatively medium level of bias, and our discussion aims to explain why certain evaluation criteria have higher or lower biases in relation to this paradigm.
However, we can also explain the bias of correctness by the paradigm. 
Correctness has the feature of extrinsic criteria since it assesses if the output sentence explains the input correctly. 
At the same time, it has the feature of intrinsic criteria since it also assesses if the output sentence provides a reasonable explanation. 
Thus, we can explain why correctness has medium bias because it is the middle of intrinsic and extrinsic criteria.

\section{Visualization and Case Study} \label{sec:app_vis}
\begin{figure}[t]
    \centering
    \begin{minipage}[b]{0.23\textwidth}
        \includegraphics[width=1.0\columnwidth]{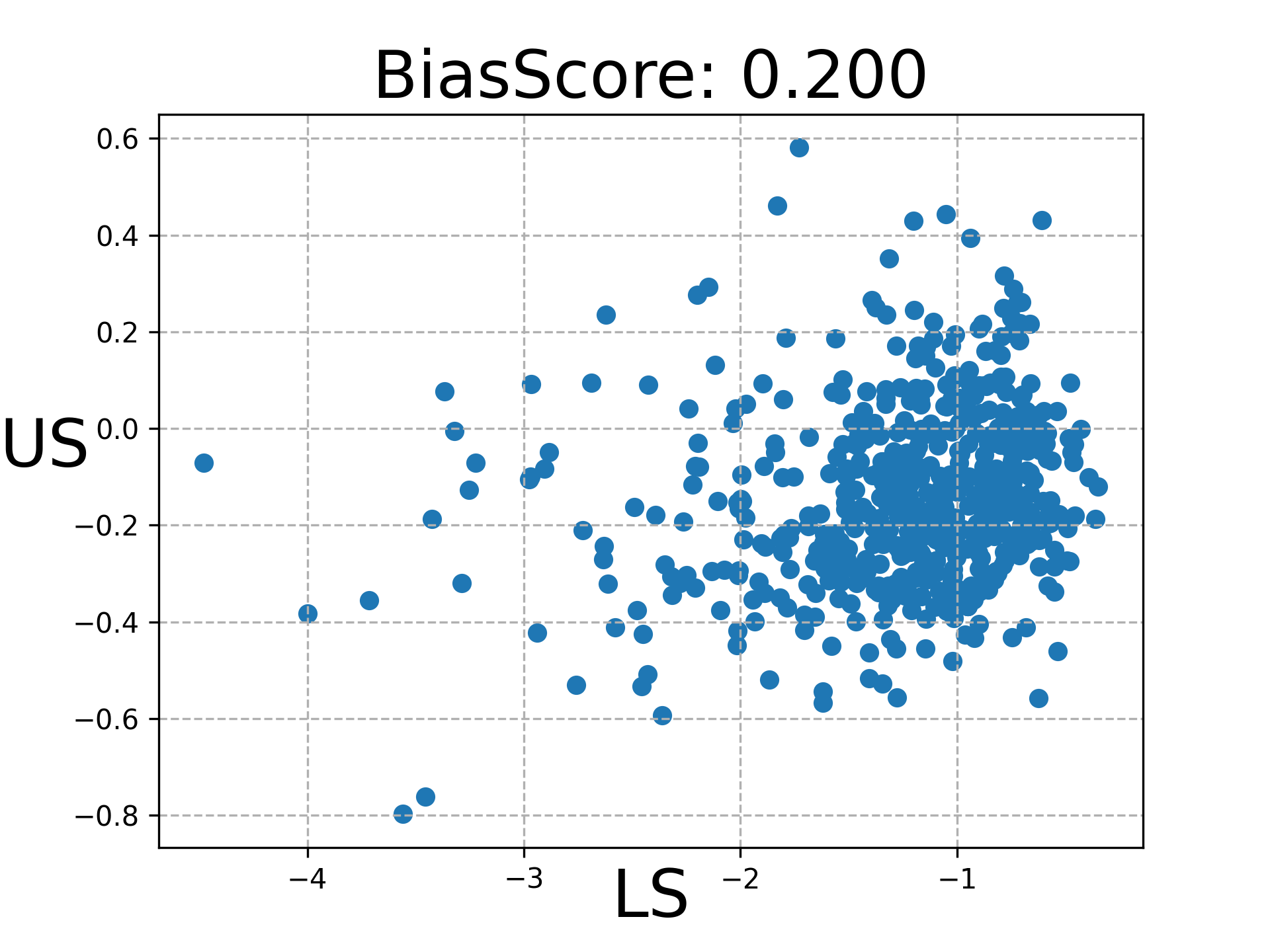}
        \subcaption{Before bias mitigation}
        \label{fig:before_mitigation}
    \end{minipage}
    \begin{minipage}[b]{0.23\textwidth}
        \includegraphics[width=1.0\columnwidth]{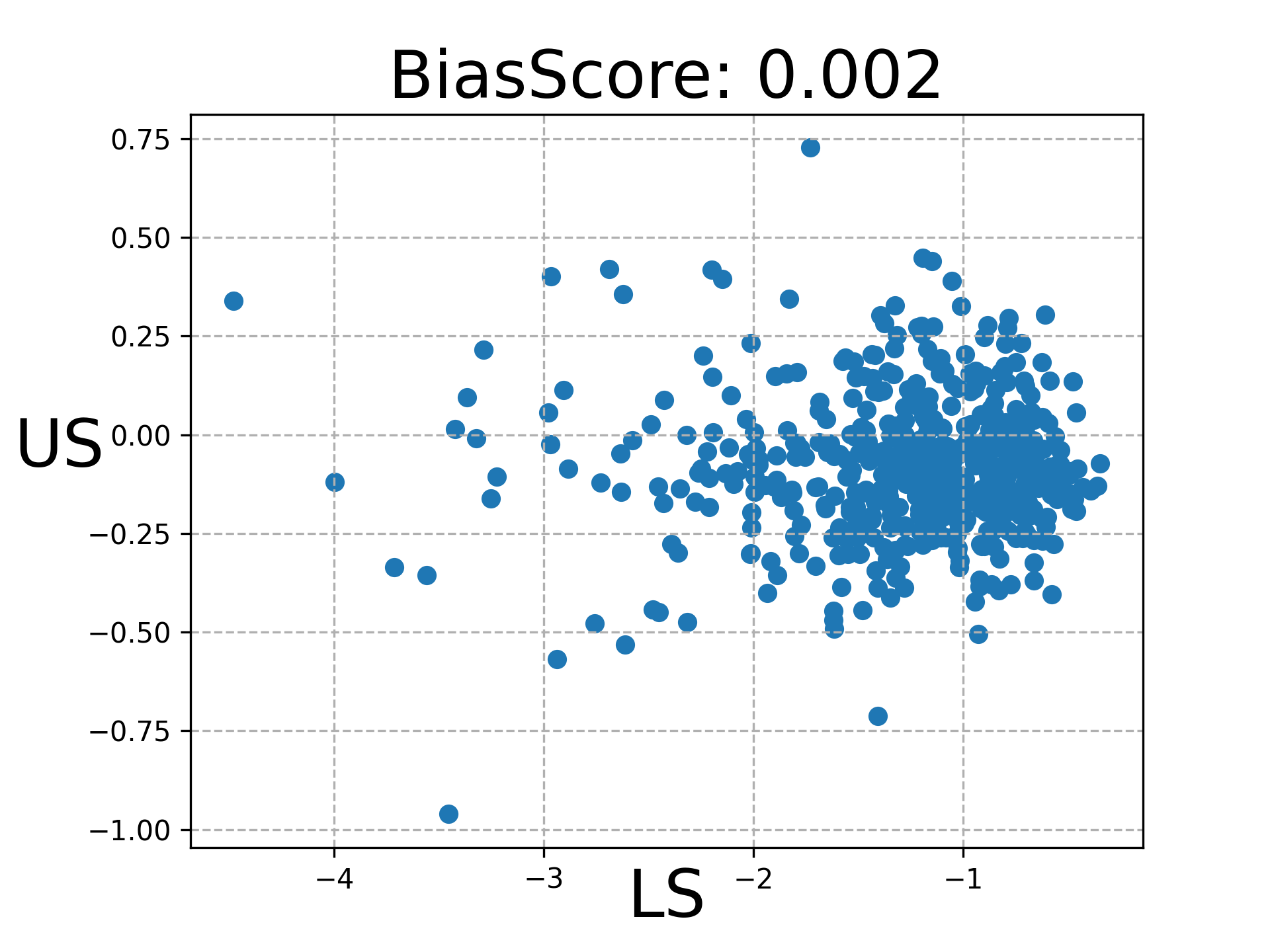}
        \subcaption{After bias mitigation}
        \label{fig:after_mitigation}
    \end{minipage}
    \caption{Visualization of the bias mitigation in Llama2-13B with data-to-text, fluency}
\end{figure}

Figures \ref{fig:before_mitigation} and \ref{fig:after_mitigation} show the visualization of likelihood bias before and after mitigation in Llama2 13B for data-to-text and fluency, respectively.
We can see that our method brings BiasScore closer to zero (0.20 to 0.00), and points are gathered to the line of US = 0, similar to (B) in Figure \ref{fig:bias_graph}.
This indicates that our method successfully mitigates likelihood bias as expected.

Below, we show an instance selected from the scatterplot where the bias has been mitigated.

\begin{itemize}
    \item Input (excerpt): (MotorSport\_Vision, city, Fawkham)
    \item Output: The Motor sport of Vision is in Fawkham.
    \item The likelihood of the output: 541st out of 568 instances in our test data.
    \item Score by humans ($\text{Score}_{\text{h}}$): 85 / 100 
    \item Score by LLM ($\text{Score}_{\text{m}}$) before bias mitigation: 2.46 / 5
    \item Score by LLM ($\text{Score}_{\text{m}}$) after bias mitigation: 4.32 / 5
\end{itemize}

Its evaluation score by humans ($\text{Score}_{\text{h}}$) is 85 out of 100, probably caused by a minor problem: the space between \textit{Motor} and \textit{sport}.
However, LLM before bias mitigation scores the instance 2.46 out of 5, which is far from $\text{Score}_{\text{h}}$.
Considering this underestimation and its low likelihood calculated by LLM (541st out of 568 instances in our test data), the score has been likely affected by likelihood bias.
After bias mitigation, LLM increased the evaluation score to 4.32, which is closer to $\text{Score}_{\text{h}}$.
These results indicate that our method successfully mitigates the bias in this instance, thus bringing the score by LLM closer to that of humans.

\end{document}